# Real-time Whole-body Obstacle Avoidance for 7-DOF Redundant Manipulators

Dake Zheng[1, 2], Xinyu Wu[1] and Jianxin Pang[2]
(dake.zheng@ubtrobot.com, xy.wu@siat.ac.cn, and walton@ubtrobot.com)

*Abstract*— **Mainly because of the heavy computational costs, the real-time whole-body obstacle avoidance for the redundant manipulators has not been well implemented. This paper presents an approach that can ensure that the whole-body of a redundant manipulator can avoid moving obstacles in real-time during the execution of a task. The manipulator is divided into end-effector and non-end-effector portion. Based on dynamical systems (DS), the real-time end-effector obstacle avoidance is obtained. Besides, the end-effector can reach the given target. By using null-space velocity control, the real-time non-end-effector obstacle avoidance is achieved. Finally, a controller is designed to ensure the whole-body obstacle avoidance. We validate the effectiveness of the method in the simulations and experiments on the 7-DOF arm of the UBTECH humanoid robot.**

## I. INTRODUCTION

Serial redundant manipulators have been widely applied in express delivery industry, manufacture industry, etc, because of their dexterity. In practice, the manipulators always have to work with other objects and agents, e.g. operators and other manipulators. We can call the agents and objects obstacles, which may be stationary or moving. In many cases, the whole body of the manipulators are expected not to collide with the obstacles during the execution of a task, e.g. pick-and-place task. As stated above, the real-time obstacle avoidance is always required by the redundant manipulators.

In order to solve the problem of obstacle avoidance in robotic systems, many methods have been proposed. Most of the approaches can be divided into reactive motion generation methods and path planning methods. Path planning methods such as those proposed in [1-3] avoid obstacles by using planning algorithms. However, the computational costs of the approaches are very heavy, those approaches can hardly be used to avoid obstacles in real-time.

Unlike the path planning approaches, the reactive motion generation approaches are proposed to achieve real-time obstacle avoidance. The methods such as the vector field histogram [4] and the curvature-velocity method [5] are able to avoid obstacles rapidly. However, those approaches are usually locally optimal, a feasible path is not always guaranteed by them. The artificial potential field method is proposed by Khatib [6]. The method is extended out by several scholars, the attractor dynamics approach is proposed by Iossifidis and Schöner [7], the dynamic potential field methods is proposed by Park et al. [8], etc. The artificial potential field method models each obstacle with a repulsive force to avoid the obstacle. The repulsive force should be well defined to avoid the local minima. To skip the local minima, the elastic band approach [9, 10] is proposed to get a collision-free path by combining the reactive techniques with the path planning algorithms. The harmonic potential methods [11, 12] are proposed and widely used [13]. Similar to the harmonic potential functions method, recently, the dynamical system (DS) based method is presented in [14]. Those methods are inspired by the description of the dynamics of fluids around impenetrable obstacles. In face of obstacles, the DS based method offers a modulation matrix of the obstacles, then the original DS is deformed by the matrix and a new path is computed according to the deformed DS to avoid the obstacles. However, it is restricted to convex obstacles. Huber et al. [15] extend out the approach and propose a method to avoid concave obstacles. However, the approach is restricted to linear DS. Since the DS is always nonlinear, the approach has only limited applications. Some other DS based obstacle avoidance approaches are proposed in [16] by using sensor-based representations of the obstacles, e.g. point cloud representation.

The obstacle avoidance for redundant manipulators has been thoroughly studied [17]. Many methods are proposed by using null-space velocity control. They assign a velocity component in the direction away from the obstacle to the point on the manipulator closest to the obstacle [18-20]. Furthermore, the task-priority approaches [17, 21, 22] are proposed. The methods ensure that the primary task, e.g. the obstacle avoidance, is executed only when needed. However, since the methods usually assume that the global path of the end-effector of the manipulator is fixed and the obstacles are always stationary, these methods may not be able to implement on more complex tasks, e.g. grab an object around multiple moving obstacles.

As stated above, although there are lots of state-of-the-art obstacle avoidance methods that have been proposed, few methods can ensure that the whole body of a redundant manipulator can avoid moving obstacles in real-time during the execution of a task. Some of the methods, e.g. the methods proposed in [1-10], can only be applied to obstacle avoidance of the end-effector of the manipulator, while some of the methods such the methods presented in [17-20] can only avoid obstacles for the non-end-effector portion of the manipulator, i.e. the rest of the manipulator except for the end-effector. Besides, the obstacles are always assumed to be stationary. To solve the real-time whole-body obstacle avoidance (RWOA) problem, this paper will present an approach based on DS and null-space velocity control. Since the 7-DOF redundant manipulators are widely used, this paper focuses on the 7-DOF redundant manipulators.

[1] Shenzhen Institutes of Advanced Technology, Chinese Academy of Sciences, Shenzhen 518055 P.R.China.
[2] UBTECH Robotics, Inc., Shenzhen 518055 P.R.China.

## II. SCHEMATIC DIAGRAM OF THE OBSTACLE AVOIDANCE APPROACH

The schematic diagram of the proposed RWOA approach is illustrated in Fig. 1.

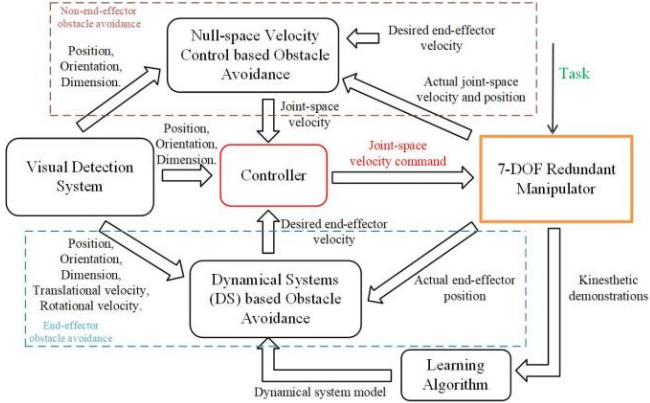

Figure 1. Schematic diagram of the RWOA approach.

As shown in Fig.1, in order to achieve whole-body obstacle avoidance, the proposed approach consists of the obstacle avoidance for the end-effector and non-end-effector. The information of obstacles such as position, orientation, velocity and dimension can be detected by the visual detection system. We can teach the manipulator to perform the motion of the given task and collect a set of kinesthetic demonstrations. Based on the demonstrations, the original DS can be estimated using the learning algorithms, e.g. the SEDS method proposed in [23]. Then, the trajectory of the end-effector of the manipulator can be computed according to the DS. With the original DS, the information of obstacles and the actual end-effector position, a new trajectory of the end-effector that can avoid the obstacles in real-time can be computed according to the DS based method proposed in [25]. With the desired end-effector velocity from the trajectory, the information of obstacles and the actual joint-space velocity and position, the non-end-effector portion of the manipulator can avoid the obstacles in real-time based on the null-space velocity control. Finally, with the information of obstacles and the obstacle avoidance algorithms for the end-effector and the non-end-effector, a controller can be synthesized to ensure real-time whole-body obstacle avoidance (RWOA) for the 7-DOF redundant manipulators. We describe the RWOA approach in details in the following sections.

## III. END-EFFECTOR OBSTACLE AVOIDANCE ALGORITHM

In this paper, we suppose there are $N$ disconnected moving convex obstacles around the manipulator. Since the manipulators usually work in the Cartesian-space, the obstacles considered are three-dimensional. In the case of connected obstacles, the obstacles can be modeled as a single convex obstacle [14]. Usually there are some non-convex objects, e.g. brushes and lamps, the Bounding Volume (BV) [24] can be used to generate three-dimensional convex envelopes for those obstacles.

For the end-effector of the manipulator, its position can be defined by $\xi$, which is three-dimensional. In the absence of obstacles, we can compute the DS based trajectory for the end-effector according to:

$$\dot{\xi} = f(\xi), \quad f: \Re^3 \mapsto \Re^3 \quad (1)$$

where $f(\cdot)$ is a continuous function. In practice, given the starting position $\{\xi\}_0$, the trajectory can be computed along time according to:

$$\{\xi\}_t = \{\xi\}_{t-1} + f(\{\xi\}_{t-1})dt \quad (2)$$

where $t$ is a positive integer and $dt$ is the integration time step.

As stated above, when there are obstacles in the original trajectory, to avoid the obstacles in real-time, a new trajectory for the end-effector should be computed immediately. According to [14], base on the original DS given by Eq.(1), we can compute the new trajectory according to:

$$\dot{\xi} = \overline{M}(\xi)(f(\xi) - \dot{\overline{\xi}}^c) + \dot{\overline{\xi}}^c \quad (3)$$

where $\overline{M}(\xi)$ is the combined modulation matrix of the obstacles, $\dot{\overline{\xi}}^c$ is the net velocity shift due to the motion of the obstacles. Eq.(3) will be deduced in the following sections.

### A. Stationary Obstacles

First, we assume the $N$ obstacles are stationary. For the $k$-th ($k = 1 \cdots N$) obstacle, $\xi_k^c$ represents the obstacle center, $\tilde{\xi}_k = \xi - \xi_k^c$ and $\Gamma_k(\tilde{\xi}_k): \Re^3 \mapsto \Re$ is a continuous distance function. The function $\Gamma_k(\tilde{\xi}_k)$ can divide the obstacle space into exterior region, boundary region and interior region as follows:

$$\begin{cases} \chi_k^e = \{\xi \in \Re^3 : \Gamma_k(\tilde{\xi}_k) > 1\} \\ \chi_k^b = \{\xi \in \Re^3 : \Gamma_k(\tilde{\xi}_k) = 1\} \\ \chi_k^i = \{\xi \in \Re^3 : \Gamma_k(\tilde{\xi}_k) < 1\} \end{cases} \quad (4)$$

where $\chi_k^e$, $\chi_k^b$ and $\chi_k^i$ are points in the exterior region, the boundary region and the interior region, respectively. Function $\Gamma_k(\tilde{\xi}_k)$ has first order partial derivatives and increases monotonically with $\|\tilde{\xi}_k\|$. $\|\tilde{\xi}_k\|$ is the distance from the $k$-th obstacle center $\xi_k^c$ to $\xi$.

In this paper, the function $\Gamma_k(\tilde{\xi}_k)$ is defined as:

$$\Gamma_k(\tilde{\xi}_k) = \sum_{i=1}^{3}((\tilde{\xi}_k)_i / a_i)^{2p_i^k} \quad (5)$$

where $(\cdot)_i$ is denoted as the $i$-th element value of a vector $(\cdot)$, $a_i$ is the $i$-th axis length, $p_i^k$ is a positive integer.

With Eq.(1), Eq.(4) and Eq.(5), we can get the deflection hyper-plane at each point $\xi$ in the exterior region of the $k$-th obstacle with normal:

$$n_k(\tilde{\xi}_k) = \begin{bmatrix} \frac{\partial \Gamma_k(\tilde{\xi}_k)}{\partial(\xi)_1} & \frac{\partial \Gamma_k(\tilde{\xi}_k)}{\partial(\xi)_2} & \frac{\partial \Gamma_k(\tilde{\xi}_k)}{\partial(\xi)_3} \end{bmatrix}^T \quad (6)$$

A linear combination of a set of two linearly independent vectors can describe every point on the hyper-plane. Here, a set of vectors consists of $e_k^1$ and $e_k^2$ is chosen as:

$$\begin{cases} e_k^1(\tilde{\xi}_k) = \begin{bmatrix} (n_k(\tilde{\xi}_k))_2 & -(n_k(\tilde{\xi}_k))_1 & 0 \end{bmatrix}^T \\ e_k^2(\tilde{\xi}_k) = \begin{bmatrix} (n_k(\tilde{\xi}_k))_3 & 0 & -(n_k(\tilde{\xi}_k))_1 \end{bmatrix}^T \end{cases} \quad (7)$$

With Eq.(5) ~ Eq.(7), define $\eta_k$ and $\sigma_k$ as the safety margin coefficient and reactivity coefficient, respectively. Define $^\eta\tilde{\xi}_k = \tilde{\xi}_k / \eta_k$. Then, the modulation matrix $M_k(^\eta\tilde{\xi}_k)$ of the $k$-th obstacle is given by:

$$\begin{cases} M_k(^\eta\tilde{\xi}_k) = E_k(^\eta\tilde{\xi}_k) D_k(^\eta\tilde{\xi}_k) E_k(^\eta\tilde{\xi}_k)^{(-1)} \\ D_k(^\eta\tilde{\xi}_k) = \begin{bmatrix} \lambda_k^1(^\eta\tilde{\xi}_k) & 0 & 0 \\ 0 & \lambda_k^2(^\eta\tilde{\xi}_k) & 0 \\ 0 & 0 & \lambda_k^3(^\eta\tilde{\xi}_k) \end{bmatrix} \\ E_k(^\eta\tilde{\xi}_k) = \begin{bmatrix} n_k(^\eta\tilde{\xi}_k) & e_k^1(^\eta\tilde{\xi}_k) & e_k^2(^\eta\tilde{\xi}_k) \end{bmatrix} \end{cases} \quad (8)$$

where $E_k(^\eta\tilde{\xi}_k)$ is a basis matrix, $D_k(^\eta\tilde{\xi}_k)$ is an associated eigenvalue matrix, $\lambda_k^1(^\eta\tilde{\xi}_k) = \begin{cases} 1 - \dfrac{\omega_k(^\eta\tilde{\xi}_k)}{|\Gamma_k(^\eta\tilde{\xi}_k)|^{\frac{1}{\sigma_k}}} & n_k(^\eta\tilde{\xi}_k)^T \dot{\xi} < 0 \\ 1 & n_k(^\eta\tilde{\xi}_k)^T \dot{\xi} \geq 0 \end{cases}$,

$\lambda_k^2(^\eta\tilde{\xi}_k) = \lambda_k^3(^\eta\tilde{\xi}_k) = 1 + \dfrac{1}{|\Gamma_k(^\eta\tilde{\xi}_k)|^{\frac{1}{\sigma_k}}}$, $\omega_k(^\eta\tilde{\xi}_k)$ is a weighting coefficient and $n_k(^\eta\tilde{\xi}_k)^T \dot{\xi}$ is used to remedy the tail-effect.

The detailed descriptions about the safety margin, the reactivity and the tail-effect can be found in [14].

Since there are $N$ obstacles, to ensure the impenetrability of the $N$ obstacles, the weighting coefficient $\omega_k(^\eta\tilde{\xi}_k)$ in Eq.(8) is given by:

$$\omega_k(^\eta\tilde{\xi}_k) = \prod_{i=1, i \neq k}^{N} \frac{(\Gamma_i(^\eta\tilde{\xi}_i)-1)}{(\Gamma_k(^\eta\tilde{\xi}_k)-1)+(\Gamma_i(^\eta\tilde{\xi}_i)-1)} \quad (9)$$

With Eq.(8) and Eq.(9), the combined modulation matrix $\overline{M}(\xi)$ of the $N$ obstacles in Eq.(3) can be formulated as:

$$\overline{M}(\xi) = \prod_{k=1}^{N} M_k(^\eta\tilde{\xi}_k) \quad (10)$$

In this section, the obstacles are assumed to be stationary. Based on the combined modulation matrix of the stationary obstacles, the effect of translational and/or rotational velocity of the moving obstacles will be discussed in the following section.

### B. Moving Obstacles

Consider $N$ moving obstacles, we denote $^L\dot{\xi}_k^c$ and $^R\dot{\xi}_k^c$ as the translational and rotational velocity of the $k$-th obstacle, respectively. To ensure that the end-effector dose not collide with any of the $N$ moving obstacles, and to ensure the smoothness of obstacle-avoidance, with Eq.(5)~Eq.(9), the net velocity shift $\bar{\dot{\xi}}^c$ in Eq.(3) can be defined as [25]:

$$\bar{\dot{\xi}}^c = \sum_{k=1}^{K} e^{-\frac{1}{\rho_k}(\Gamma_k(^\eta\tilde{\xi}_k)-1)} \omega_k(^\eta\tilde{\xi}_k)(^L\dot{\xi}_k^c + {}^R\dot{\xi}_k^c \times {}^\eta\tilde{\xi}_k) \quad (11)$$

where $\rho_k$ is a positive coefficient for tuning the smoothing term $e^{-\frac{1}{\rho_k}(\Gamma_k(^\eta\tilde{\xi}_k)-1)}$, $(\cdot) \times (\cdot)$ is denoted as the cross product.

With Eq.(10) and Eq.(11), according to [14, 25], in the presence of $N$ moving obstacles, the new trajectory of the end-effector computed according to Eq.(3) can avoid the obstacles in real-time. Besides, the new trajectory can reach the same target as the original trajectory computed according to Eq.(1). Therefore, if the end-effector can track the new trajectory, it can avoid the obstacles and eventually reach the given target.

## IV. NON-END-EFFECTOR OBSTACLE AVOIDANCE ALGORITHM

So far we have shown that the end-effector of the manipulator can avoid the obstacles and reach the given target by tracking the trajectory computed according to Eq.(3). However, it is easy to believe that when the end-effector tracks the trajectory the rest of the manipulator except for the end-effector, i.e. the non-end-effector, will collide with the obstacles. To get the obstacle avoidance for the non-end-effector while maintaining the obstacle avoidance for the end-effector, the null-space velocity control method will be used.

### A. Preliminaries

The 7-DOF manipulator is considered in this paper, generally, the dimension of its task space $m$ is 6 and the dimension of its joint space $n$ is 7, therefore, the degree of redundancy $r$, i.e. the difference between $n$ and $m$, is 1. If $m$ is less than 6 then the manipulator has a high degree of redundancy.

We denote $q$ as the variable of joint position and $x$ as task variable of the end-effector. Then, the relationship between $q$ and $x$ can be described as [26]:

$$x = g(q) \quad (12)$$

where $g(\cdot)$ is a 6-dimensional function vector.

Differentiating Eq.(12), we can get:

$$\dot{x} = J\dot{q} \quad (13)$$

where $J$ is a $6 \times 7$ Jacobian matrix.

In general, the control problem is how to get the joint velocity $\dot{q}$ that will result in the desired end-effector velocity $\dot{x}$. According to Eq.(13), since the Jacobian matrix $J$ is not a square matrix, given the desired end-effector velocity $\dot{x}$, the joint velocity $\dot{q}$ can be calculated by:

$$\dot{q} = J^{\#}\dot{x} + N\dot{\varphi} \quad (14)$$

where $J^{\#}$ is the generalized inverse of the Jacobian matrix $J$, $N$ is a matrix denoted as the projection into the null-space of the matrix $J$, $\dot{\varphi}$ is a 7-dimensional joint velocity vector.

Usually, the norm of the joint velocity $\dot{q}$ should be minimum, therefore the Moor-Penrose inverse [27] $J^+$, $J^+ = J^T(JJ^T)^{(-1)}$ is chosen as the generalized inverse $J^{\#}$. Then Eq.(14) can be further written as:

$$\dot{q} = \underbrace{J^+}_{J^{\#}} \dot{x} + \underbrace{(I - J^+ J)}_{N} \dot{\varphi} \quad (15)$$

where the term $J^+\dot{x}$ ensures the minimum norm of joint velocity $\dot{q}$. The term $(I-J^+J)\dot{\varphi}$ offers different sets of joint velocity $\dot{q}$ that result in the same end-effector velocity $\dot{x}$. With the term $(I-J^+J)\dot{\varphi}$, the obstacle avoidance for the non-end-effector portion of the manipulator can be obtained. The definition of the velocity $\dot{\varphi}$ is presented in the following sections.

### B. Null-space Velocity Control based Obstacle Avoidance

Fig.2 illustrates the the obstacle avoidance strategy using the null-space velocity control. Given Eq.(15), the strategy is to identify the point $C_0$ on the non-end-effector portion closest to the obstacles and then apply velocity $\dot{x}_0$ at that point if the point, or critical point, is within the critical distance $d_m$ of the obstacles. The vector $d_0$ goes from the closest point on the obstacles to point $C_0$ to point $C_0$, and the direction of $\dot{x}_0$ is the same as the direction of $d_0$ [18, 21]. In addition, $J_0$ is denoted as the Jacobian matrix from the base to point $C_0$.

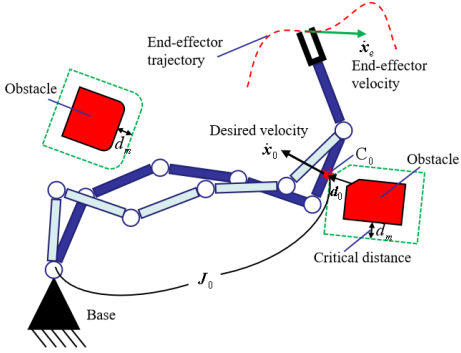

Figure 2. Obstacle avoidance strategy using null-space velocity control.

According to Eq.(13), the end-effector velocity $\dot{x}_e$ and the critical point $C_0$ velocity $\dot{x}_0$ can be described by:
$$\dot{x}_e = J\dot{q}, \quad \dot{x}_0 = J_0\dot{q} \tag{16}$$
Substituting $\dot{x}$ in Eq.(14) with $\dot{x}_e$ given by Eq.(16), we get:
$$\dot{\varphi} = (J_0 N)^{\#}(\dot{x}_0 - J_0 J^{\#}\dot{x}_e) \tag{17}$$
Substituting $\dot{\varphi}$ in Eq.(14) with Eq.(17), we obtain:
$$\dot{q} = J^{\#}\dot{x} + N(J_0 N)^{\#}(\dot{x}_0 - J_0 J^{\#}\dot{x}_e) \tag{18}$$
Since $N$ in Eq.(15) is idempotent and hermitian [18], we have $(J_0 N)^{\#} = N(J_0 N)^{\#}$. Then Eq.(18) can be rewritten as:
$$\dot{q} = J^{\#}\dot{x} + (J_0 N)^{\#}(\dot{x}_0 - J_0 J^{\#}\dot{x}_e) \tag{19}$$
The first r.h.s term in Eq.(19) can be used to design a controller that allows the end-effector to accurately track the given trajectory. For the controller, $\dot{x}$ is used instead of $\dot{x}_e$. The term $J_0 J^{\#}\dot{x}_e$ is denoted as the velocity at point $C_0$ caused by the end-effector velocity. The term $(J_0 N)^{\#}$ transforms the given critical point velocity $\dot{x}_0$ in the task space into the joint space.

In practice, $\dot{x}_0$ is always 3-dimensional, i.e. $\dot{x}_0 = [vx_0 \ vy_0 \ vz_0]$, $vx_0$, $vy_0$ and $vz_0$ are the velocities along the $x$-axis, $y$-axis and $z$-axis, respectively. Therefore we can only obtain the exact velocity $\dot{x}_0$ if the degree of redundancy $r$ of the manipulator is more than 3. However the degree of redundancy of the 7-DOF manipulator is always 1. To avoid this problem, the reduced operational space method [28] will be used. As shown in Fig.2, we have:
$$\dot{x}_0 = v_0 n_0, \quad n_0 = \frac{d_0}{\|d_0\|} \tag{20}$$
where $n_0$ is the unit vector of vector $d_0$. Define $J_{d0} = n_0^T J_0$, then Eq.(19) can be rewritten as:
$$\dot{q} = J^{\#}\dot{x} + (J_{d0} N)^{\#}(\dot{x}_0 - J_{d0} J^{\#}\dot{x}_e) \tag{21}$$
where $v_0$ and the term $J_{d0} J^{\#}\dot{x}_e$ are scalar.

Given Eq.(21), to obtain a good obstacle avoidance performance, the scalar velocity $v_0$ should be well selected.

### C. Selection of the Critical Point Velocities

As shown in Fig.2, a good choice of the velocity $v_0$ is [21]:
$$\begin{cases} v_0 = \delta_v v_n \\ \delta_v = \begin{cases} (\frac{d_m}{\|d_0\|})^2 - 1 & \text{for } \|d_0\| < d_m \\ 0 & \text{for } \|d_0\| \geq d_m \end{cases} \end{cases} \tag{22}$$
where $v_n$ is a nominal velocity, $\delta_v$ is a coefficient.

To further smooth the motion of obstacle avoidance, a smoothing factor $\delta_h$ is presented in [18] to smoothly apply the homogenous solution, i.e. the second r.h.s term in Eq.(21) to the manipulator. Then we have:
$$\dot{q} = J^{\#}\dot{x} + \delta_h (J_{d0} N)^{\#}(v_0 - J_{d0} J^{\#}\dot{x}_e) \tag{23}$$
with
$$\delta_h = \begin{cases} 1 & \text{for } \|d_0\| \leq d_m \\ \frac{1}{2}(1+\cos(\pi \frac{\|d_0\|-d_m}{d_i-d_m})) & \text{for } d_m < \|d_0\| < d_i \\ 0 & \text{for } \|d_0\| \geq d_i \end{cases} \tag{24}$$
where $d_i$ is the distance at which the obstacle begins to affect the motion of the manipulator.

By choosing factors $\delta_v$ and $\delta_h$, we can get a obstacle avoidance motion for the non-end-effector of the manipulator with a good trade-off between the speed and the smoothness.

## V. CONTROLLER

So far we have obtained the end-effector obstacle avoidance and the non-end-effector obstacle avoidance. Base on the obtained results, a proper controller should be designed to ensure the whole body obstacle avoidance for the manipulator. Fig.3 illustrates the end-effector tracking the trajectory computed according to Eq.(3).

As shown in Fig. 3, $E_0$ is denoted as the end-effector center, $\dot{x}_d = [(\dot{\xi})_1 \ (\dot{\xi})_2 \ (\dot{\xi})_3 \ \omega x \ \omega y \ \omega z]$ represents the desired end-effector velocity including the translational velocity given by Eq.(3) and the rotational velocity given by the desired rotational positions, $\omega x$, $\omega y$ and $\omega z$ are the given rotational velocities along $x$-axis, $y$-axis and $z$-axis, respectively. $d_s$ represents the safety distance of the obstacle determined by the safety margin coefficient in Eq.(8), the vector $d_e$ goes from the closest point on the obstacle to center $E_0$ to center $E_0$. According to [14], part of the trajectory always overlap with the safety boundary of the obstacle. However, due to the trajectory tracking error, the center $E_0$ will penetrate the safety boundary. Therefore, the trajectory cannot be computed according to Eq.(3).

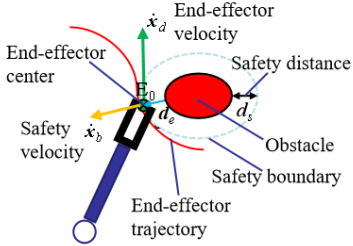

Figure 3. Illustration of the end-effector tracking the trajectory.

To avoid the problem, the safety velocity $\dot{x}_b$ in the direction of vector $d_e$ is added to the desired velocity $\dot{x}_d$ as shown in Fig.3. To ensure the smoothness of the end-effector motion, the velocity $\dot{x}_b$ is defined as:

$$\begin{cases} \dot{x}_b = \varsigma_v v_b \dfrac{d_e}{\|d_e\|} \\ \varsigma_v = \begin{cases} 1 & \text{for } \|d_e\| \le d_s \\ \dfrac{1}{2}(1+\cos(\pi \dfrac{\|d_e\|-d_s}{d_k-d_s})) & \text{for } d_s < \|d_e\| < d_k \\ 0 & \text{for } \|d_e\| \ge d_k \end{cases} \end{cases} \quad (25)$$

where $v_b$ is a nominal safety velocity, $\varsigma_v$ is a smoothing factor similar to Eq.(24), $d_k$ is the distance at which the obstacle begins affect the motion of the end-effector. Since the safety velocity $\dot{x}_b$ is translational, the safety velocity with fixed zero rotational velocities can be defined as

$$\dot{x}_{br} = [(\dot{x}_b)_1 \ (\dot{x}_b)_2 \ (\dot{x}_b)_3 \ 0 \ 0 \ 0] \quad (26)$$

As described in Fig. 3, with Eq.(2), Eq.(3) and Eq.(26), $\dot{x}_e$ and $\dot{x}$ in Eq.(23) can be defined as:

$$\dot{x}_e = \dot{x}_d + \dot{x}_{br} \quad (27)$$
$$\dot{x} = \dot{x}_e + K_a(\dot{x}_e dt - \xi_r) \quad (28)$$

where $\xi_r$ is the actual position and orientation of the end-effector, $K_a$ is a control gain matrix which is a $6\times 6$ diagonal matrix, the term $(\dot{x}_e dt - \xi_r)$ is the trajectory tracking error of the end-effector. With a properly chosen matrix $K_a$, the trajectory tracking stability and accuracy will be guaranteed [21, 26].

With Eq.(27) and Eq.(28), the desired control commands for the joints of the manipulator can be computed by Eq.(23). Finally, refer to Fig.1, the control commands can ensure the RWOA for the 7-DOF manipulator.

## VI. SIMULATION AND EXPERIMENT VALIDATION

To validate the effectiveness of the proposed RWOA approach, we implement the proposed method and the obstacle avoidance approach proposed in [25] on the 7-DOF arm of the UBTECH humanoid robot, and carry out several sets of comparative simulation and experiment. In the simulations and experiments the arm is controlled at a rate of 100 Hz. Since we don't have a visual detection system at present, we assume that the position, orientation and velocity of the obstacles are known. In the experiments we move the obstacles by hand approximately at a given velocity.

Fig.4(a) shows the simulation set-up in the ROS plus Gazebo environment and Fig.4(b) shows the corresponding experiment set-up. For simplicity, only the left arm and the torso of the humanoid robot are shown. There are two moving ellipsoidal obstacles around the arm. In the simulations and experiments obstacle 1 is moving along the $x$-axis at the velocity of 1 m/s and obstacle 3 is moving along the $x$-axis at the velocity of 2 m/s. Refer to Fig.2, the purple line represents the line from the closest point on the obstacles to the critical point on the non-end-effector portion of the arm to the critical point, and the green line represents the trajectory of the end-effector.

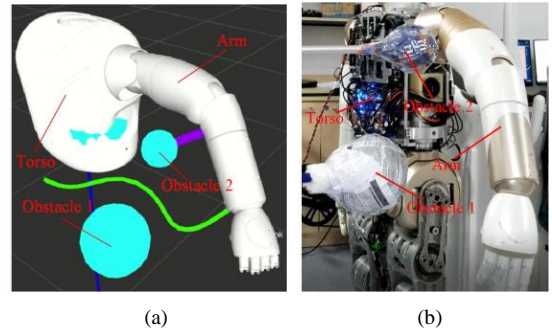

Figure 4. Validation set-up.

In the simulations and experiments, in the absence of obstacles, an original trajectory computed according to a given original DS is given for the end-effector. In the presence of the two moving obstacles, we carry out three sets of simulation and experiment.

(1) When we ignore the presence of obstacles, and the end-effector simply tracks the original trajectory, the actual position and orientation and the desired position and orientation of the end-effector are converted to joint velocities using the task space controller given in [17] and the inverted kinematics of the arm. The simulation and the corresponding experiment results show that the end-effector has collision with obstacle 1 and obstacle 2 collides with the non-end-effector portion of the arm. That makes sense, because we didn't implement any obstacle avoidance algorithms on the arm.

(2) When the arm is under the control of the DS based method proposed in [25], while the non-end-effector obstacle avoidance is ignored, the simulation and the corresponding experiment results show that the end-effector avoids obstacle 1 while the non-end-effector portion of the arm collides with obstacle 2. As described in [25], in the presence of moving obstacles, the original trajectory is deformed according to DS method to avoid the obstacle in real-time, therefore the end-effector can avoid the obstacle 1 by tracking the deformed trajectory. However, the DS based method cannot ensure obstacle avoidance for the non-end-effector portion of the arm, hence the non-end-effector cannot avoid obstacle 2.

(3) When the arm is under the control of the proposed RWOA approach, the simulation and the corresponding experiment results show that the whole body of the arm has no collisions with the obstacles and the end-effector reaches at the given target finally. During the validation, the trajectory of the end-effector is computed in real-time to avoid the obstacles. Besides the purple line indicates that the non-end-effector portion of the arm is avoiding the obstacle by using the null-space velocity control.

As stated above, the simulation and the corresponding results indicate the effectiveness of the proposed RWOA approach.

## VII. CONCLUSION

In this paper, we presented a RWOA approach for the 7-DOF redundant manipulators. The original DS was deformed by the combined modulation matrix of the moving obstacles, then the trajectory of the end-effector was computed according to the deformed DS that ensure that the end-effector tracking it can avoid the obstacles in real-time and reach the desired target. When the end-effector tracks the trajectory, the null-space velocity control was used to ensure the real-time obstacle avoidance for the non-end-effector portion. Finally, a Cartesian pose feedback controller with safety velocity compensation that can prevent the end-effector from penetrating the obstacles was designed. We implemented the method on the 7-DOF arm of the UBTECH humanoid robot. The simulation and experiment results indicate the effectiveness of the RWOA approach. The proposed approach is general and can be applied to redundant manipulators with other degrees of freedom.